\documentclass[runningheads]{llncs}
\usepackage{graphicx}
\usepackage{comment}
\usepackage{amsmath,amssymb} 
\usepackage{color}
\usepackage{subfigure}
\usepackage{multirow}
\usepackage{threeparttable}
\usepackage[symbol]{footmisc}
\begin{document}
\renewcommand{\thefootnote}{\fnsymbol{footnote}}
\pagestyle{headings}
\mainmatter
\def\ECCVSubNumber{3822}  

\title{Attention-Driven Dynamic Graph Convolutional Network for Multi-Label Image Recognition} 

\titlerunning{Attention-Driven Dynamic GCN Network for Multi-Label Image Recognition}
%
\author{Jin Ye\inst{1*} \and
Junjun He\inst{1,2*} \and
Xiaojiang Peng\inst{1}\thanks{Equally-contributed first authors. $^\dagger$Corresponding author (yu.qiao@siat.ac.cn)} \and
Wenhao Wu\inst{1} \and
Yu Qiao\inst{1}$^\dagger$}
%
%
\institute{ShenZhen Key Lab of Computer Vision and Pattern Recognition, Shenzhen Institutes of Advanced Technology, Chinese Academy of Sciences, Shenzhen, China. \and
School of Biomedical Engineering, the Institute of Medical Robotics, Shanghai Jiao Tong University, Shanghai, China.
}
\maketitle

\begin{abstract}
Recent studies often exploit Graph Convolutional Network (GCN) to model label dependencies to improve recognition accuracy for multi-label image recognition. However, constructing a graph by counting the label co-occurrence possibilities of the training data may degrade model generalizability, especially when there exist occasional co-occurrence objects in test images. Our goal is to eliminate such bias and enhance the robustness of the learnt features. To this end, we propose an Attention-Driven Dynamic Graph Convolutional Network (ADD-GCN) to dynamically generate a specific graph for each image. ADD-GCN adopts a Dynamic Graph Convolutional Network (D-GCN) to model the relation of content-aware category representations that are generated by a Semantic Attention Module (SAM). Extensive experiments on public multi-label benchmarks demonstrate the effectiveness of our method, which achieves mAPs of 85.2\%, 96.0\%, and 95.5\% on MS-COCO, VOC2007, and VOC2012, respectively, and outperforms current state-of-the-art methods with a clear margin. All codes can be found at \url{https://github.com/Yejin0111/ADD-GCN}.

\end{abstract}

\section{Introduction}
Nature scenes usually contains multiple objects. In the computer vision community, multi-label image recognition is a fundamental computer vision task and plays a critical role in wide applications such as human attribute recognition~\cite{li2016human}, medical image recognition~\cite{ge2018chest} and recommendation systems~\cite{jain2016extreme,yang2015pinterest}. Unlike single-label classification, multi-label image recognition needs to assign multiple labels to a single image. Therefore it is reasonable to take account of the relationships of different labels to enhance recognition performance. 

    \begin{figure}[t]
    \begin{center}
    \subfigure[Example of static graph]{
    \includegraphics[width=0.42\columnwidth]{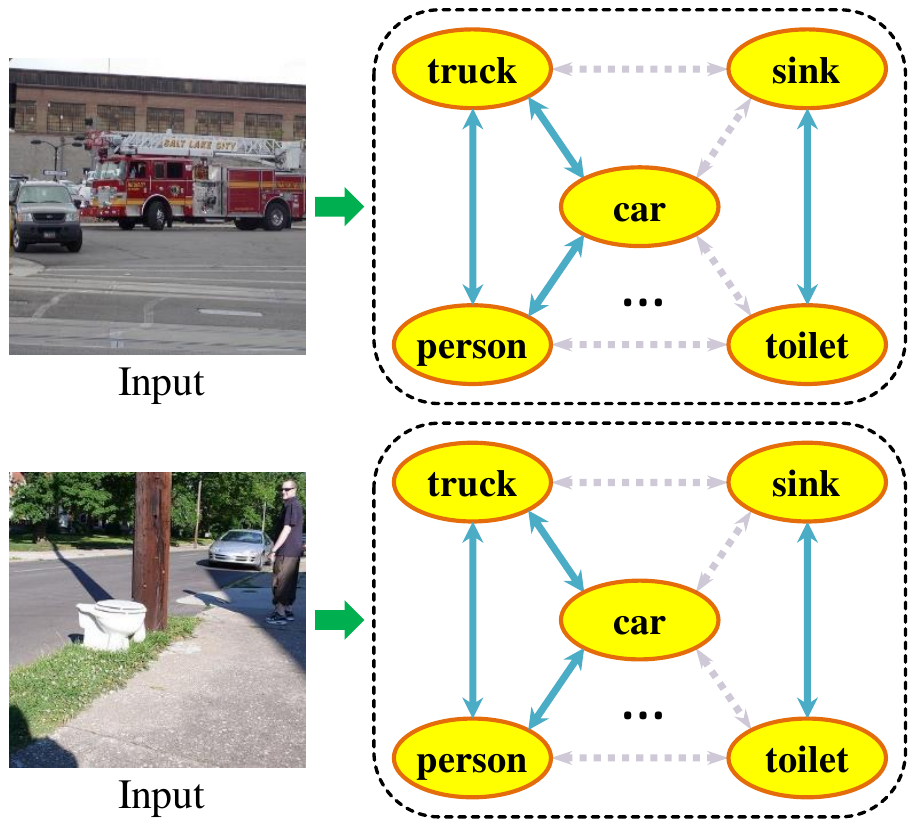}
    \label{fig:motivation-static}
    }
    \hfill
    \subfigure[Example of dynamic graph]{
    \includegraphics[width=0.42\columnwidth]{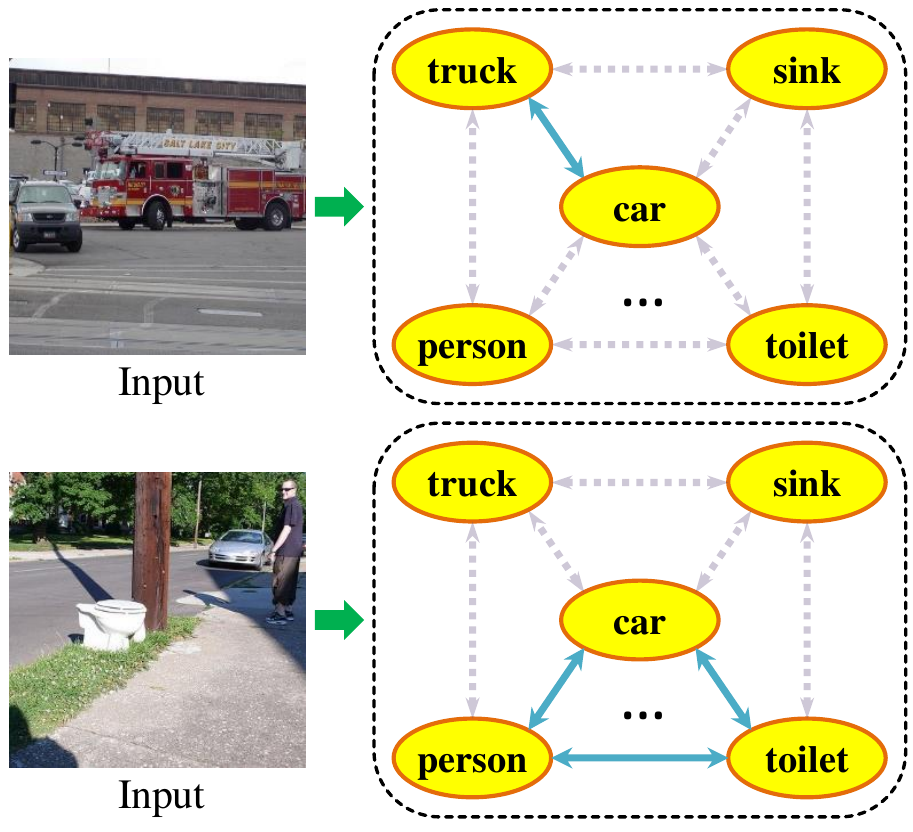}
    \label{fig:motivation-dynamic}
    }
    \caption{Static graph and dynamic graph. Solid line indicates higher relation and dashed line indicates lower relation of the categories. (a) illustrates all images share a static graph~\cite{chen2019learning,chen2019graph}. (b) shows our motivation that different image has its own graph that can describe the relations of co-occurred categories in the image.}
    \label{fig:motivation}
    \end{center}
    \end{figure}
    
Recently, Graph Convolutional Network (GCN)~\cite{kipf2017semi} achieves great success in modeling relationship among vertices of a graph. Current state-of-the art methods~\cite{chen2019learning,chen2019multi} build a complete graph to model the label correlations between each two categories by utilizing prior frequency of label co-occurrence of the target dataset and achieved remarkable results. However, building such a global graph for the whole dataset could cause the frequency-bias problem in most common datasets. As highlighted in~\cite{tommasi2017deeper,torralba2011unbiased}, most prominent vision datasets are afflicted with the co-occur frequency biases despite the best efforts of their creators. Let us consider a common category ``car'', which always appears with different kind of vehicles such as ``truck'', ``motorbike'', and ``bus''. This may inadvertently cause the frequency-bias in these datasets, which would guide the model to learn higher relations among them. Specifically, as shown in Fig~\ref{fig:motivation-static}, each image share a static graph which is built by calculating the co-occurrence frequency of categories in target dataset. The static graph gives higher relation values between ``car'' and ``truck'' and lower ones between ``car'' and ``toilet''in each image. This may result in several problems as follows: 1) failing to identify ``car'' in a different context such as in the absence of ``truck'', 2) hallucinating ``truck'' even in a scene containing only ``car'', and 3) ignoring ``toilet'' when ``car'' co-occurs with ``toilet''. 

Given these issues, our goal is to build a dynamic graph that can capture the content-aware category relations for each image. Specifically, as shown in Fig~\ref{fig:motivation-dynamic}, we construct the image-specific dynamic graph in which ``car'' and ``toilet'' has strong connections for the image that ``car'' and ``toilet'' appear together and vice versa. To this end, we propose a novel Attention-Driven Dynamic Graph Convolutional Network (ADD-GCN) for multi-label image recognition which leverages content-aware category representations to construct dynamic graph representation. Unlike previous graph based methods~\cite{chen2019learning,chen2019multi}, 
ADD-GCN models semantic relation for each input image by estimating an image-specific dynamic graph. Specifically, we first decompose the convolutional feature map into multiple content-aware category representations through the Semantic Attention Module (SAM). Then we feed these representations into a Dynamic GCN (D-GCN) module which performs feature propagation via two joint graphs: static graph and dynamic graph. Finally discriminative vectors are generated by D-GCN for multi-label classification. The static graph mainly captures coarse label dependencies over the training dataset and learns such semantic relations as shown in Fig~\ref{fig:motivation-static}. The correlation matrix of dynamic graph is the output feature map of a light-weight network applied upon content-aware category representations for each image, and is used to capture fine dependencies of those content-aware category representations as illustrated in Fig~\ref{fig:motivation-dynamic}.

Our main contributions can be summarized as follows,
\begin{itemize}
\item The major contribution of this paper is that we introduce a novel dynamic graph constructed from content-aware category representations for multi-label image recognition. The dynamic graph is able to capture category relations for a specific image in an adaptive way, which further enhance its representative and discriminative ability.
\item We elaborately design an end-to-end Attention-Driven Dynamic Graph Convolutional Network (ADD-GCN), which consists of two joint modules. i) Semantic Attention Module (SAM) for locating semantic regions and producing content-aware category representations for each image, and ii) Dynamic Graph Convolutional Network (D-GCN) for modeling the relation of content-aware category representations for final classification. 
\item Our ADD-GCN significantly outperforms recent state-of-the-art approaches on popular multi-label datasets: MS-COCO, VOC2007, and VOC2012. Specifically, our ADD-GCN achieves mAPs of 85.2\% on MS-COCO, 96.0\% on VOC2007, and 95.5\% on VOC2012, respectively, which are new records on these benchmarks.
\end{itemize}

\section{Related work}
Recent renaissance of deep neural network remarkably accelerates the progresses in single-label image recognition. Convolutional Neural Networks (CNNs) can learn powerful features from large scale image datasets such as MS-COCO~\cite{lin2014microsoft}, PASCAL VOC~\cite{everingham2010pascal} and ImageNet~\cite{deng2009imagenet}, which greatly alleviates the difficulty of designing hand-crafted features.   
Recently, many CNN-based approaches have been proposed for multi-label image recognition as well~\cite{chen2019learning,cheng2014bing,girshick2015fast,ren2015faster,zitnick2014edge,wang2016cnn,wang2017multi}, which can be roughly categorized into two main directions as following.

\textbf{Region based methods}. One direction aims to first coarsely localize multiple regions and then recognize each region with CNNs~\cite{cheng2014bing,girshick2015fast,ren2015faster,zitnick2014edge}. Wei~\textit{et al.}~\cite{wei2015hcp} propose a Hypotheses-CNN-Pooling (HCP) framework which generates a large number of proposals by objectness detection methods~\cite{cheng2014bing,zitnick2014edge} and treats each proposal as a single-label image recognition problem. Yang~\textit{et al.}~\cite{yang2016exploit} formulate the task as a multi-class multi-instance learning problem. Specifically, they incorporate local information by generating a bag of instances for each image and enhance the discriminative features with label information. However, these object proposal based methods lead to numerous category-agnostic regions, which make the whole framework sophisticated and require massive computational cost. Moreover, these methods largely ignore the label dependencies and region relations, which are essential for multi-label image recognition. 
        
\textbf{Relation based methods}. Another direction aims to exploit the label dependencies or region relations~\cite{wang2016cnn,wang2017multi,li2016conditional,chen2019multi,chen2019learning,wang2019multi}.
Wang~\textit{et al.}~\cite{wang2016cnn} propose CNN-RNN framework to predict the final scores and formulate label relation by utilizing Recurrent Neural Network (RNN). Wang~\textit{et al.}~\cite{wang2017multi} attempt to discover such relations by iterative locating attention regions with spatial transformer~\cite{jaderberg2015spatial} and LSTM~\cite{hochreiter1997long}. Actually, these RNN/LSTM based methods explore the relation between labels or semantic regions in a sequential way, which cannot fully exploit the direct relations among them. Different from these sequential methods, some works resort to Graphical architectures. Li~\textit{et al.}~\cite{li2016conditional} cope with such relations by image-dependent conditional label structures with Graphical Lasso framework. Li~\textit{et al.}~\cite{Li:2014:MIC:3020751.3020796} use a maximum spanning tree algorithm to create a tree-structured graph in the label space. Recently, the remarkable capacity of Graph Convolutional Networks (GCNs) has been proved in several vision tasks, Chen~\textit{et al.}~\cite{chen2019multi} utilize GCN to propagate prior label representations (e.g. word embeddings) and generate a classifier, which replaces the last linear layer in a normal deep convolutional neural network such as ResNet~\cite{he2016deep}. With the help of label annotations, Chen~\textit{et al.}~\cite{chen2019learning} compute a probabilistic matrix as the relation edge between each label in a graph.
Our work is largely inspired by these GCN based methods for multi-label image recognition. However, instead of using external word embedding for  category representations and label statistics for graph construction, our Attention-Driven Dynamic Graph Convolutional Network (ADD-GCN) directly decomposes the feature map extracted by a CNN backbone into content-aware category representations and optimizes the D-GCN, which consists of a static graph for capturing the global coarse category dependencies and a dynamic graph for exploiting content-dependent category relations, respectively.

    \begin{figure}
    \begin{center}
        \includegraphics[width=0.95\linewidth]{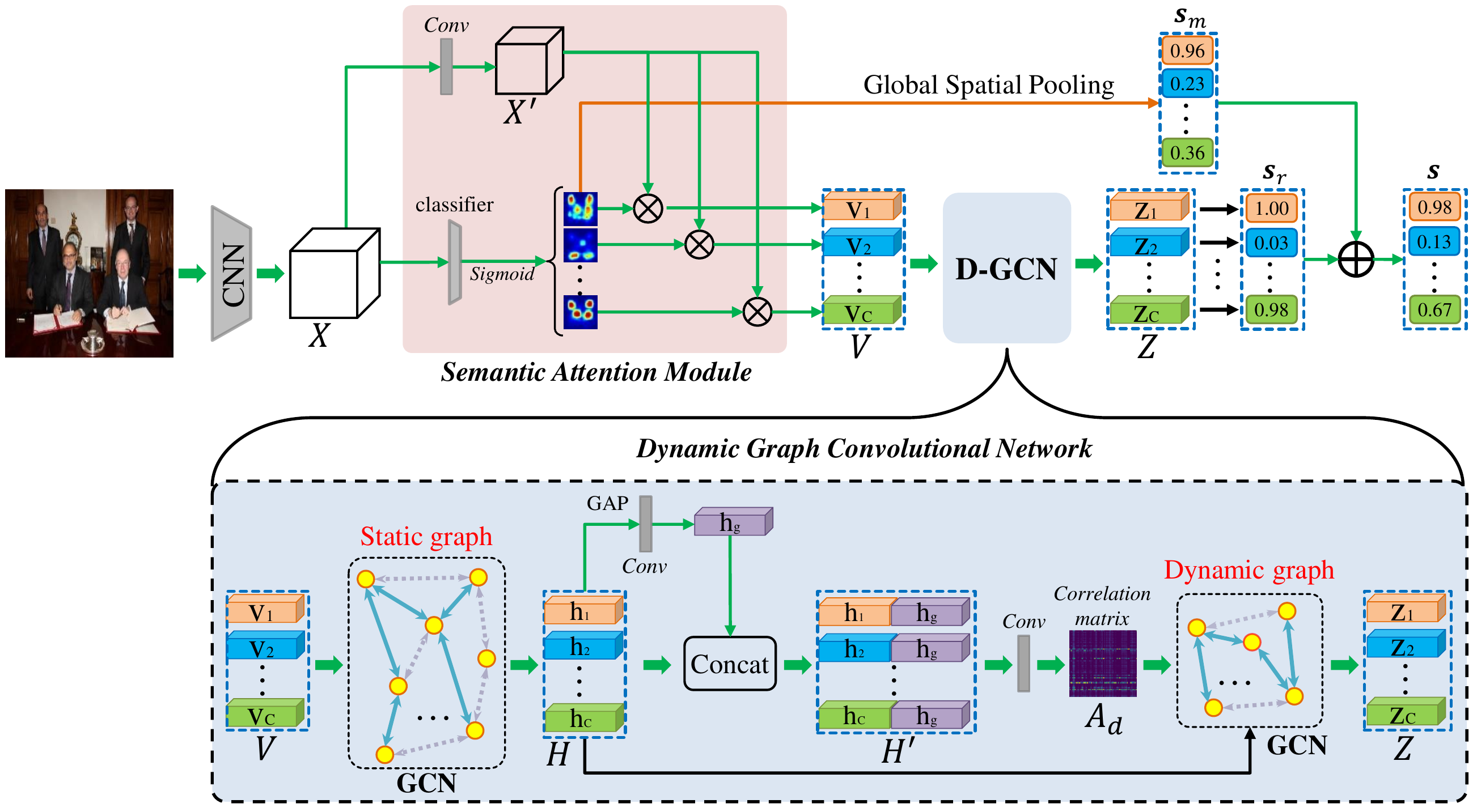}
    \end{center}
    \caption{Overall framework of our approach. Given an image, ADD-GCN first uses a CNN backbone to extract convolutional feature maps $\textbf{X}$. Then, SAM decouples $\textbf{X}$ to content-aware category representations $\textbf{V}$, and D-GCN models global and local relations among $\textbf{V}$ to generate the final robust representations $\textbf{Z}$ that contains rich relation information with other categories.
    }
    \label{fig:framework}
    \end{figure}
\section{Method}
This section presents Attention-Driven Dynamic Graph Convolutional Network (ADD-GCN) for multi-label image recognition. We first give a brief overview of ADD-GCN, and then describe its key modules (Semantic Attention Module and Dynamic GCN module) in details.
    
\subsection{Overview of ADD-GCN}
As objects always co-occur in image, how to effectively capture the relations among them is important for multi-label recognition. Graph based representations provide a practical way to model label dependencies. We can use nodes $\textbf{V}=[\textbf{v}_1,\textbf{v}_2,\dots,\textbf{v}_C]$ to represent labels and correlation matrix $\textbf{A}$ to represent the label relations (edges). Recent studies~\cite{chen2019learning,chen2019multi} exploited Graph Convolutional Network (GCN) to improve the performance of multi-label image recognition with a clear margin. However, they construct correlation matrix $\textbf{A}$ in a static way, which mainly accounts for the label co-occurrence in the training dataset, and is fixed for each input image. As a result, they fail to explicitly utilize the content of each specific input image.

To address this problem, this paper proposes ADD-GCN with two elaborately designed modules: We first introduce Semantic Attention Module (SAM) to estimate content-aware category representation $\textbf{v}_c$ for each class $c$ from the extracted feature map and the representations are input to another module, Dynamic-GCN, for final classification. We will detail them in the next part. 
    
\subsection{Semantic Attention Module}\label{sec:SAM}
The objective of Semantic Attention Module (SAM) is to obtain a set of content-aware category representations, each of which describes the contents related to a specific label from input feature map $\textbf{X}\in\mathbb{R}^{{H}\times{W}\times{D}}$. As shown in Fig~\ref{fig:framework}, SAM first calculates category-specific activation maps $\textbf{M}=[\textbf{m}_1,\textbf{m}_2,\dots,\textbf{m}_C]\in\mathbb{R}^{{H}\times{W}\times{C}}$ and then they are used to convert the transformed feature map $\textbf{X}'\in\mathbb{R}^{{H}\times{W}\times{D'}}$ into the content-aware category representations $\textbf{V}=[\textbf{v}_1,\textbf{v}_2,\dots,\textbf{v}_C]\in\mathbb{R}^{C\times D}$ . Specifically, each class representation $\textbf{v}_c$ is formulated as a weighted sum on $\textbf{X}'$ as follows, such that the produced $\textbf{v}_c$ can selectively aggregate features related to its specific category $c$.
\begin{equation}
    \textbf{v}_c=\textbf{m}_c^T\textbf{X}'=\sum_{i=1}^{H}\sum_{j=1}^{W}{m_{i,j}^{c}\textbf{x}'_{i,j}},
    \label{eq:semantic-location}
\end{equation}
where $m_{i,j}^{c}$ and $\textbf{x}'_{i,j}\in\mathbb{R}^{D'}$ are the weight of \textit{c}~th activation map and the feature vector of the feature map at $(i,j)$, respectively. Then the problem reduces to how to calculate the category-specific activation maps $\textbf{M}$, where the difficulty comes from we do not have explicit supervision like bounding box or category segmentation for images. 

\textbf{Activation map generation}.    
We generate the category-specific activation maps $\textbf{M}$ based on Class Activation Mapping (CAM)~\cite{zhou2016cvpr}, which is a technique to expose the implicit attention on an image without bounding box and segmentation. 
Specifically, we can perform Global Average Pooling (GAP) or Global Max Pooling (GMP) on the feature map $\textbf{X}$ and classify these pooled features with FC classifiers. Then these classifiers are used to identify the category-specific activation maps by convolving the weights of FC classifiers with feature map $\textbf{X}$. Unlike CAM, we put a convolution layer as the classifier as well as a $Sigmoid(\cdot)$ to regularize $\textbf{M}$ before the global spatial pooling, which has better performance in experiments. Ablation studies on these methods are presented in Table~\ref{tab:ablation-spatial-pool}.
    
\subsection{Dynamic GCN}
With content-aware category representations $\textbf{V}$ obtained in previous section, we introduce Dynamic GCN (D-GCN) to adaptively transform their coherent correlation for multi-label recognition. Recently Graph Convolutional Network (GCN)~\cite{kipf2017semi} has been widely proven to be effective in several computer vision tasks and is applied to model label dependencies for multi-label image recognition with a static graph~\cite{chen2019learning,chen2019multi}. Different from these works, we propose a novel D-GCN to fully exploit relations between content-aware category representations to generate discriminative vectors for final classification. Specifically, our D-GCN consists of two graph representations, static graph and dynamic graph, as shown in Fig~\ref{fig:framework}. We first revisit the traditional GCN and then detail our D-GCN. 
    
\textbf{Revisit GCN}. Given a set of features $\textbf{V}\in\mathbb{R}^{{C}\times{D}}$ as input nodes, GCN aims to utilize a correlation matrix $\textbf{A}\in\mathbb{R}^{{C}\times{C}}$ and a state-update weight matrix $\textbf{W}\in\mathbb{R}^{{D}\times{D_u}}$ to update the values of $\textbf{V}$. Formally, The updated nodes $\textbf{V}_{u}\in\mathbb{R}^{{C}\times{D_u}}$ can be formulated by a single-layer GCN as 
    \begin{equation}
        \textbf{V}_{u} = \delta(\textbf{A}\textbf{V}\textbf{W}),
        \label{eq:gcn-review}
    \end{equation}
where $\textbf{A}$ is usually pre-defined and $\textbf{W}$ is learned during training. $\delta(\cdot)$ denotes an activation function, such as the $ReLU(\cdot)$ or $Sigmoid(\cdot)$, which makes the whole operation nonlinear. The correlation matrix $\textbf{A}$ reflects the relations between the features of each node. During inference, the correlation matrix $\textbf{A}$ first diffuse the correlated information among all nodes, then each node receives all necessary information and its state is updated through a linear transformation $\textbf{W}$.
    
\textbf{D-GCN}. As shown in the bottom of Fig~\ref{fig:framework}, D-GCN takes the content-aware category representations $\textbf{V}$ as input node features, and sequentially feeds them into a static GCN and a dynamic GCN. Specifically, the single-layer static GCN is simply defined as $\textbf{H} = LReLU(\textbf{A}_s\textbf{V}\textbf{W}_s)$, where $\textbf{H}=[\textbf{h}_1,\textbf{h}_2,\dots,\textbf{h}_C]\in\mathbb{R}^{{C}\times{D_1}}$, the activation function $LReLU(\cdot)$ is LeakyReLU, and the correlation matrix $\textbf{A}_s$ and state-update weights $\textbf{W}$ is randomly initialized and learned by gradient decent during training. Since $\textbf{A}_s$ is shared for all images, it is expected that $\textbf{A}_s$ can capture global coarse category dependencies.

In the next, we introduces dynamic GCN to transform $\textbf{H}$, whose correlation matrix $\textbf{A}_d$ is estimated adaptively from input features $\textbf{H}$. Note this is different from static GCN whose correlation matrix is fixed and shared for all input samples after training, while our $\textbf{A}_d$ is constructed dynamically dependent on input feature. Since every sample has different $\textbf{A}_d$, it makes model increase its representative ability and decrease the over-fitting risk that static graph brings. Formally, the output $\textbf{Z}\in\mathbb{R}^{C\times{D_2}}$ of the dynamic GCN can be defined as, 
\begin{equation}
\textbf{Z} = f(\textbf{A}_d\textbf{H}\textbf{W}_d), ~ \textrm{where}~
        \textbf{A}_d = \delta(\textbf{W}_A\textbf{H}'),
        \label{eq:Ad-generation}
\end{equation}
where $f(\cdot)$ is the LeakyReLU activation function, $\delta(\cdot)$ is the Sigmoid activation function, $\textbf{W}_d\in \mathbb{R}^{{D}_1\times{D}_2}$ is the state-update weights, $\textbf{W}_A\in \mathbb{R}^{C\times{2D_1}}$ is the weights of a \textit{conv} layer to formulate the dynamic correlation matrix $\textbf{A}_d$, and $\textbf{H}'\in\mathbb{R}^{{2D_1} \times C }$ is obtained by concatenating $\textbf{H}$ and its global representations $\textbf{h}_g\in\mathbb{R}^{D_1}$, which is obtained by global average pooling and one \textit{conv} layer, sequentially. Formally, $\textbf{H}'$ is defined as, 
    \begin{equation}
        \textbf{H}' = [(\textbf{h}_{1};\textbf{h}_g),(\textbf{h}_{2};\textbf{h}_g),\dots,(\textbf{h}_{c};\textbf{h}_g)].
        \label{eq:concat-A2-Ag}
    \end{equation}
It is worth mentioning that the dynamic graph $\textbf{A}_d$ is specific for each image which may capture content-dependent category dependencies. Overall, our D-GCN enhances the content-aware category representations from $\textbf{V}$ to $\textbf{Z}$ by the dataset-specific graph and the image-specific graph.
    
\subsection{Final Classification and Loss}
\textbf{Final Classification}. As shown in Fig~\ref{fig:framework}, the final category representation $\textbf{Z}=[\textbf{z}_1,\textbf{z}_2,\dots,\textbf{z}_C]$ is used for final classification. Due to each vector $\textbf{z}_i$ is aligned with its specific class and contains rich relation information with others, we simply put each category vector into a binary classifier to predict its category score. In particular, we concatenate the score for each category to generate the final score vector $\textbf{s}_r=[s_r^1,s_r^2,\dots,s_r^C]$. In addition, we can also get another confident scores $\textbf{s}_m=[s_m^1,s_m^2,\dots,s_m^C]$ through global spatial pooling on the category-specific activation map $\textbf{M}$ estimated by SAM in Section~\ref{sec:SAM}. Thus, we can aggregate the two score vectors to predict more reliable results. Here we simply average them to produce the final scores $\textbf{s} = [s^1,s^2,\dots,s^C]$.
 
 \textbf{Training Loss}. We supervise the final score $\textbf{s}$ and train the whole ADD-GCN with the traditional multi-label classification loss as follows,
\begin{equation}
    L(\textbf{y},\textbf{s}) = \sum_{c=1}^{C}y^{c}\log(\sigma(s^{c}))+(1-y^{c})\log(1-\sigma(s^{c})),
    \label{eq:cr-loss}
\end{equation}
where $\sigma(\cdot)$ is $Sigmoid(\cdot)$ function.

\section{Experiments}
In this section, we first introduce the evaluation metrics and our implementation details. And then, we compare our ADD-GCN with other existing state-of-the-art methods on three public multi-label image recognition dataset, \textit{i.e.}, MS-COCO~\cite{lin2014microsoft}, Pascal VOC 2007~\cite{everingham2010pascal}, and Pascal VOC 2012~\cite{everingham2010pascal}. Finally, we conduct extensive ablation studies and present some visualization results of the category-specific activation maps and the dynamic graphs.

\subsection{Evaluation Metrics}
To compare with other existing methods in a fair way, we follow previous works~\cite{chen2019learning,chen2019multi,zhu2017learning} to adopt the average of overall/per-class precision (OP/CP), overall/per-class recall (OR/CR), overall/per-class F1-score (OF1/CF1) and the mean Average Precision (mAP) as evaluation metrics. When measuring precision/recall/F1-score, the label is considered as positive if its confident score is great than 0.5. Besides, top-3 results of precision/recall/F1-score are also reported. Generally, the OF1, CF1 and mAP are more important than other metrics.

\subsection{Implementation Details} 
For the whole ADD-GCN framework, we use ResNet-101~\cite{he2016deep} as our backbone. The channel of $\textbf{V}$ is 1024 and the nonlinear activation function LeakyReLU with negative slop of 0.2 is adopted in our SAM and D-GCN. During training, we adopt the data augmentation suggested in~\cite{chen2019multi} to avoid over-fitting: the input image is random cropped and resized to $448\times448$ with random horizontal flips for data augmentation. To make our model converge quickly, we follow~\cite{chen2019learning} to choose the model that trained on COCO as the pre-train model for Pascal VOC. We choose SGD as our optimizer, with momentum of 0.9 and weight decay of $10^{-4}$. The batch size of each GPU is 18. The initial learning rate is set to 0.5 for SAM/D-GCN and 0.05 for backbone CNN. We train our model for 50 epoch in total and the learning rate is reduced by a factor of 0.1 at 30 and 40 epoch, respectively. All experiments are implemented based on PyTorch~\cite{paszke2017automatic}.

\subsection{Comparison with State of The Arts}
To demonstrate the scalability and effectiveness of our proposed ADD-GCN, extensive
experiments are conducted on three widely used benchmarks, i.e., MS-COCO~\cite{lin2014microsoft}, Pascal VOC 2007~\cite{everingham2010pascal}, and Pascal VOC 2012~\cite{everingham2010pascal}.

\begin{table*}
\caption{Comparison of our ADD-GCN and other state-of-the-art methods on MS-COCO dataset. The best results are marked as bold.}
\begin{center}
\label{tab:coco-result}
\resizebox{\textwidth}{!}{ %
\begin{threeparttable}
\begin{tabular}{|l|c|c|c|c|c|c|c|c|c|c|c|c|c|}
\hline
\multirow{2}{*}{Method} & \multicolumn{7}{|c|}{All} & \multicolumn{6}{|c|}{Top-3} \\
\cline{2-14}
~ & mAP & CP & CR & CF1 & OP & OR & OF1 & CP & CR & CF1 & OP & OR & OF1 \\
\hline\hline
RARL~\cite{chen2018recurrent}       &    - &    - &    - &    - &    - &    - &    - & 78.8 & 57.2 & 66.2 & 84.0 & 61.6 & 71.1 \\
RDAR~\cite{wang2017multi}           &    - &    - &    - &    - &    - &    - &    - & 79.1 & 58.7 & 67.4 & 84.0 & 63.0 & 72.0 \\
Multi-Evidence~\cite{ge2018multi}   &    - & 80.4 & 70.2 & 74.9 & 85.2 & 72.5 & 78.4 & 84.5 & 62.2 & 70.6 & 89.1 & 64.3 & 74.7 \\
ResNet-101~\cite{he2016deep}        & 79.7 & 82.7 & 67.4 & 74.3 & 86.4 & 71.8 & 78.4 & 85.9 & 60.5 & 71.0 & 90.2 & 64.2 & 75.0 \\
DecoupleNet~\cite{guo2019multi}     & 82.2 & 83.1 & 71.6 & 76.3 & 84.7 & 74.8 & 79.5 &    - &    - &    - &    - &    - &    - \\
ML-GCN~\cite{chen2019multi}         & 83.0 & 85.1 & 72.0 & 78.0 & 85.8 & 75.4 & 80.3 & 89.2 & 64.1 & 74.6 & 90.5 & 66.5 & 76.7 \\
SSGRL~\cite{chen2019learning}       & 83.8 & \textbf{89.9} & 68.5 & 76.8 & \textbf{91.3} & 70.8 & 79.7 & \textbf{91.9} & 62.5 & 72.7 & \textbf{93.8} & 64.1 & 76.2 \\
\hline\hline
Ours    & \textbf{85.2} & 84.7 & \textbf{75.9} & \textbf{80.1} & 84.9 & \textbf{79.4} & \textbf{82.0} & 88.8 & \textbf{66.2} & \textbf{75.8} & 90.3 & \textbf{68.5} & \textbf{77.9} \\ 
\hline
\end{tabular}
\end{threeparttable}}
\end{center}
\end{table*}

\textbf{MS-COCO}. Microsoft COCO~\cite{lin2014microsoft} is primarily built for object segmentation and detection, and it is also widely used for multi-label recognition recently. It is composed of a training set with 82081 images, a validation set with 40137 images. The dataset covers 80 common object categories with about 2.9 object labels per image. The number of labels for each image varies considerably, rendering MS-COCO more challenging. Since the labels of the test set are not given, we compare our performance to other previous methods on the validation set. 

Table~\ref{tab:coco-result} shows the comparison between our ADD-GCN and other state-of-the-art methods. In particular, we compare with RARL~\cite{chen2018recurrent}, RDAR~\cite{wang2017multi}, Multi-Evidence~\cite{ge2018multi}, ResNet-101~\cite{he2016deep}, DecoupleNet~\cite{guo2019multi}, ML-GCN~\cite{chen2019multi}, and SSGRL~\cite{chen2019learning}. Our ADD-GCN consistently outperforms the other state-of-the-art approaches in terms of OF1, CF1, and mAP, as well as some other less important metrics. In particular, both ML-GCN and SSGRL also construct graphs for multi-label classification, our ADD-GCN respectively outperforms ML-GCN by 2.2\% and SSGRL by 1.4\% in terms of mAP. In addition, our ADD-GCN improves the baseline by 5.5\%. This demonstrates the superiority of our approach.

\begin{table}[t]
\caption{Comparison of our ADD-GCN and other state-of-the-art methods on Pascal VOC 2007 dataset. The best results are marked as bold.}
\begin{center}
\label{tab:voc07-result}
\resizebox{\textwidth}{!}{ %
\begin{tabular}{|l|c|c|c|c|c|c|c|c|c|c|c|c|c|c|c|c|c|c|c|c||c|}
\hline
Method & aero & bike & bird & boat & bottle & bus & car & cat & chair & cow & table & dog & horse & mbike & person & plant & sheep & sofa & train & tv & mAP \\
\hline\hline
CNN-RNN~\cite{wang2016cnn}         & 96.7 & 83.1 & 94.2 & 92.8 & 61.2 & 82.1 & 89.1 & 94.2 & 64.2 & 83.6 & 70.0 & 92.4 & 91.7 & 84.2 & 93.7 & 59.8 & 93.2 & 75.3 & 99.7 & 78.6 & 84.0 \\
RMIC~\cite{he2018reinforced}            & 97.1 & 91.3 & 94.2 & 57.1 & 86.7 & 90.7 & 93.1 & 63.3 & 83.3 & 76.4 & 92.8 & 94.4 & 91.6 & 95.1 & 92.3 & 59.7 & 86.0 & 69.5 & 96.4 & 79.0 & 84.5 \\
RLSD~\cite{zhang2018multilabel}         & 96.4 & 92.7 & 93.8 & 94.1 & 71.2 & 92.5 & 94.2 & 95.7 & 74.3 & 90.0 & 74.2 & 95.4 & 96.2 & 92.1 & 97.9 & 66.9 & 93.5 & 73.7 & 97.5 & 87.6 & 88.5 \\
VeryDeep~\cite{simonyan2014very}        & 98.9 & 95.0 & 96.8 & 95.4 & 69.7 & 90.4 & 93.5 & 96.0 & 74.2 & 86.6 & 87.8 & 96.0 & 96.3 & 93.1 & 97.2 & 70.0 & 92.1 & 80.3 & 98.1 & 87.0 & 89.7 \\
ResNet-101~\cite{he2016deep}      & 99.1 & 97.3 & 96.2 & 94.7 & 68.3 & 92.9 & 95.9 & 94.6 & 77.9 & 89.9 & 85.1 & 94.7 & 96.8 & 94.3 & 98.1 & 80.8 & 93.1 & 79.1 & 98.2 & 91.1 & 90.8 \\
HCP~\cite{wei2015hcp}             & 98.6 & 97.1 & 98.0 & 95.6 & 75.3 & 94.7 & 95.8 & 97.3 & 73.1 & 90.2 & 80.0 & 97.3 & 96.1 & 94.9 & 96.3 & 78.3 & 94.7 & 76.2 & 97.9 & 91.5 & 90.9 \\
RDAR~\cite{wang2017multi}            & 98.6 & 97.4 & 96.3 & 96.2 & 75.2 & 92.4 & 96.5 & 97.1 & 76.5 & 92.0 & 87.7 & 96.8 & 97.5 & 93.8 & 98.5 & 81.6 & 93.7 & 82.8 & 98.6 & 89.3 & 91.9 \\
FeV+LV~\cite{yang2016exploit}  & 98.2 & 96.9 & 97.1 & 95.8 & 74.3 & 94.2 & 96.7 & 96.7 & 76.7 & 90.5 & 88.0 & 96.9 & 97.7 & 95.9 & 98.6 & 78.5 & 93.6 & 82.4 & 98.4 & 90.4 & 92.0 \\
RARL~\cite{chen2018recurrent}            & 98.6 & 97.1 & 97.1 & 95.5 & 75.6 & 92.8 & 96.8 & 97.3 & 78.3 & 92.2 & 87.6 & 96.9 & 96.5 & 93.6 & 98.5 & 81.6 & 93.1 & 83.2 & 98.5 & 89.3 & 92.0 \\
RCP~\cite{wang2016beyond}             & 99.3 & 97.6 & 98.0 & 96.4 & 79.3 & 93.8 & 96.6 & 97.1 & 78.0 & 88.7 & 87.1 & 97.1 & 96.3 & 95.4 & 99.1 & 82.1 & 93.6 & 82.2 & 98.4 & 92.8 & 92.5 \\
ML-GCN~\cite{chen2019multi}    & 99.5 & 98.5 & \textbf{98.6} & 98.1 & 80.8 & 94.6 & 97.2 & 98.2 & 82.3 & 95.7 & 86.4 & 98.2 & 98.4 & 96.7 & 99.0 & 84.7 & 96.7 & 84.3 & 98.9 & 93.7 & 94.0 \\
SSGRL~\cite{chen2019learning} & 99.7 & 98.4 & 98.0 & 97.6 & 85.7 & 96.2 & 98.2 & \textbf{98.8} & 82.0 & 98.1 & \textbf{89.7} & \textbf{98.8} & 98.7 & 97.0 & 99.0 & 86.9 & 98.1 & 85.8 & 99.0 & 93.7 & 95.0 \\
\hline\hline
Ours             & \textbf{99.8} & \textbf{99.0} & 98.4 & \textbf{99.0} & \textbf{86.7} & \textbf{98.1} & \textbf{98.5} & 98.3 & \textbf{85.8} & \textbf{98.3} & 88.9 & \textbf{98.8} & \textbf{99.0} & \textbf{97.4} & \textbf{99.2} & \textbf{88.3} & \textbf{98.7} & \textbf{90.7} & \textbf{99.5} & \textbf{97.0} & \textbf{96.0} \\
\hline
\end{tabular}}
\end{center}
\end{table}

\textbf{VOC 2007}. 
Pascal VOC 2007~\cite{everingham2010pascal} is widely used multi-label dataset, which contains 9963 images from 20 common object categories. It is divided into a train set, a validation set, and a test set. For fair comparisons, following previous works~\cite{chen2019learning,chen2019multi}, we train our model on the trainval set (5011 images) and evaluate on the test set (4952 images). The evaluation metrics are the Average Precision (AP) and the mean of Average Precision (mAP).

The comparison between our ADD-GCN and other methods is presented in Table~\ref{tab:voc07-result}. Our method consistently outperforms these methods with a clear margin, and improves our baseline from 90.8\% to 96.0\%. Particularly, 
compared with other two current state-of-the-art methods ML-GCN and SSGRL~\cite{chen2019learning}, the gain of overall mAP is 2.0\% and 1.0\%, respectively.

\begin{table}[t]
\caption{Comparison of our ADD-GCN and other state-of-the-art methods on Pascal VOC 2012 dataset. The best results are marked as bold.}
\begin{center}
\label{tab:voc12-result}
\resizebox{\textwidth}{!}{ %
\begin{tabular}{|l|c|c|c|c|c|c|c|c|c|c|c|c|c|c|c|c|c|c|c|c||c|}
\hline
Methods                 & aero & bike & bird & boat & bottle & bus & car & cat & chair & cow & table & dog & horse & mbike & person & plant & sheep & sofa & train & tv & mAP \\
\hline
RMIC~\cite{he2018reinforced}                    & 98.0 & 85.5 & 92.6 & 88.7 & 64.0 & 86.8 & 82.0 & 94.9 & 72.7 & 83.1 & 73.4 & 95.2 & 91.7 & 90.8 & 95.5 & 58.3 & 87.6 & 70.6 & 93.8 & 83.0 & 84.4 \\
VeryDeep~\cite{simonyan2014very}               & 99.1 & 88.7 & 95.7 & 93.9 & 73.1 & 92.1 & 84.8 & 97.7 & 79.1 & 90.7 & 83.2 & 97.3 & 96.2 & 94.3 & 96.9 & 63.4 & 93.2 & 74.6 & 97.3 & 87.9 & 89.0 \\
HCP~\cite{wei2015hcp}                     & 99.1 & 92.8 & 97.4 & 94.4 & 79.9 & 93.6 & 89.8 & 98.2 & 78.2 & 94.9 & 79.8 & 97.8 & 97.0 & 93.8 & 96.4 & 74.3 & 94.7 & 71.9 & 96.7 & 88.6 & 90.5 \\
FeV+LV~\cite{yang2016exploit}                  & 98.4 & 92.8 & 93.4 & 90.7 & 74.9 & 93.2 & 90.2 & 96.1 & 78.2 & 89.8 & 80.6 & 95.7 & 96.1 & 95.3 & 97.5 & 73.1 & 91.2 & 75.4 & 97.0 & 88.2 & 89.4 \\
RCP \cite{wang2016beyond}                     & 99.3 & 92.2 & 97.5 & 94.9 & 82.3 & 94.1 & 92.4 & 98.5 & 83.8 & 93.5 & 83.1 & 98.1 & 97.3 & 96.0 & 98.8 & 77.7 & 95.1 & 79.4 & 97.7 & 92.4 & 92.2 \\
SSGRL \cite{chen2019learning}            & 99.7 & 96.1 & 97.7 & 96.5 & 86.9 & 95.8 & 95.0 & 98.9 & 88.3 & 97.6 & 87.4 & 99.1 & \textbf{99.2} & 97.3 & 99.0 & 84.8 & 98.3 & 85.8 & 99.2 & 94.1 & 94.8 \\
\hline\hline
Ours                & \textbf{99.8} & \textbf{97.1} & \textbf{98.6} & \textbf{96.8} & \textbf{89.4} & \textbf{97.1} & \textbf{96.5} & \textbf{99.3} & \textbf{89.0} & \textbf{97.7} & \textbf{87.5} & \textbf{99.2} & 99.1 & \textbf{97.7} & \textbf{99.1} & \textbf{86.3} & \textbf{98.8} & \textbf{87.0} & \textbf{99.3} & \textbf{95.4} & \textbf{95.5} \\
\hline
\end{tabular}}
\end{center}
\end{table}

\textbf{VOC 2012}. 
Pascal VOC 2012~\cite{everingham2010pascal} is the dataset that is widely used for multi-label image recognition task, which consists of 11540 images as trainval set and 10991 as test set from 20 common object categories. For fair comparisons with previous state-of-the-art methods, we train our model on the trainval set and evaluate our results on test set.

We present the AP of each category and mAP over all categories of VOC 2012 in Table~\ref{tab:voc12-result}. Our ADD-GCN also achieves the best performance compared with other state-of-the-art methods. Concretely, the proposed ADD-GCN obtains 95.5\% mAP, which outperforms another state-of-the-art SSGRL by 0.7\%. And the AP of each category is higher than other methods except ``horse''. The results demonstrate the effectiveness of our framework. 
    
\subsection{Ablation Studies} 
    In this section, we conduct ablation experiments on MS-COCO and VOC 2007. 

    \begin{figure}
    \begin{center}
    \subfigure[Comparisons on MS-COCO.]{
    \includegraphics[width=0.47\columnwidth]{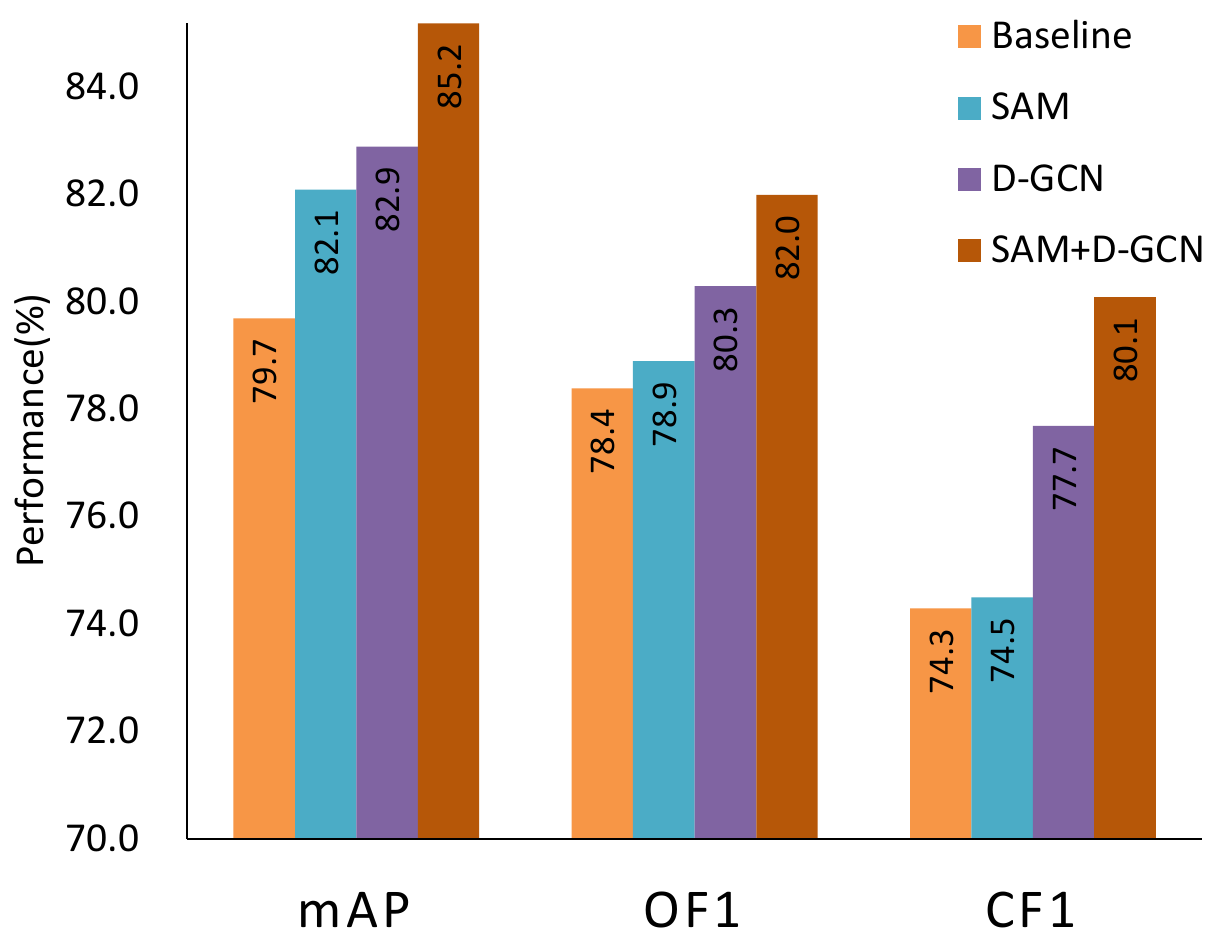}
    \label{fig:ablation-module-coco}
    }
    \hfill
    \subfigure[Comparisons on VOC 2007.]{
    \includegraphics[width=0.47\columnwidth]{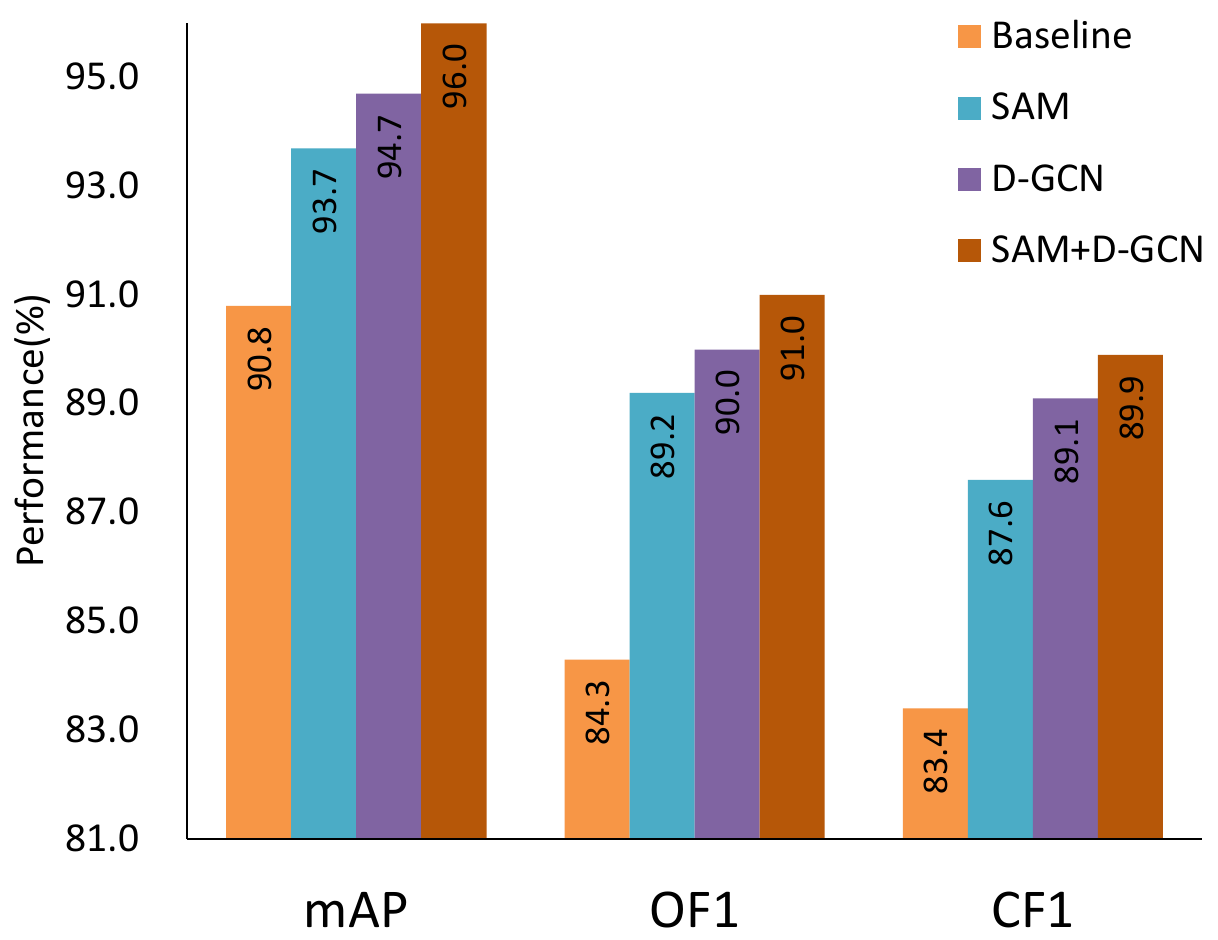}
    \label{fig:ablation-module-voc}
    }
    \end{center}
    \caption{Evaluation of SAM and D-GCN on MS-COCO and VOC 2007.}
    \label{fig:ablation-module}
    \end{figure}
\textbf{Evaluation of SAM and D-GCN}.
To investigate the contribution of each module in ADD-GCN, we separately apply SAM and D-GCN with certain adaptions upon a standard ResNet backbone. We evaluate the effectiveness of SAM by removing D-GCN and adding binary classifiers upon the output of SAM~($\textbf{V}$) directly, while for evaluating the effectiveness of D-GCN, we simply replace SAM with a \textit{Conv-LReLU} block. The results are shown in Fig~\ref{fig:ablation-module}. As can be seen, on both MS-COCO and VOC-2007, SAM and D-GCN individually improve the baseline with large margins. Compared to the baseline that directly learn classifier upon global-pooled features, SAM first decomposes the feature map into content-aware category representations and train classifiers upon them. The improvement from SAM shows that the decomposed representations are more discriminative. We also find that D-GCN is able to enhance the discriminative ability of features from the results compared with the baseline. Combining SAM and D-GCN further boosts performance as we expect, since they focus on different aspects. Specifically, the gain of the mAP, OF1 and CF1 over the baseline is 5.5\%, 3.6\% and 5.8\% on MS-COCO, while 5.2\%, 6.7\% and 6.5\% on VOC 2007.

\begin{table}[tp]
\begin{center}
\caption{The performance of different combinations of static and dynamic graph. ``S'': static graph and ``D'': dynamic graph. ``P'': we propagate information through the static and dynamic graph in a parallel way, and fuse them by either addition (add) or element-wise multiplication (mul) or concatenation (cat).}
\label{tab:ablation-combination}
\begin{tabular}{|l|c|c|c||c|c|c|}
\hline
\multirow{2}{*}{Methods} & \multicolumn{3}{|c||}{All (COCO)} & \multicolumn{3}{|c|}{All (VOC 2007)} \\
\cline{2-7}
~           & mAP  & OF1  & CF1  & mAP  & OF1  & CF1  \\
\hline\hline
ResNet-101  & 79.7 & 78.4 & 74.3 & 90.8 & 84.3 & 83.4 \\
S           & 82.9 & 78.3 & 74.7 & 94.5 & 89.3 & 88.3 \\
D           & 83.7 & 79.4 & 76.6 & 94.9 & 89.9 & 88.7 \\
P (add)     & 84.0 & 79.4 & 76.9 & 94.6 & 88.8 & 88.2 \\
P (mul)     & 83.7 & 80.8 & 78.5 & 94.6 & 89.6 & 88.5 \\
P (cat)     & 83.3 & 80.0 & 76.9 & 94.9 & 89.7 & 88.8 \\
D$\to$S     & 84.5 & 81.4 & 79.3 & 95.0 & 90.1 & 88.8 \\
S$\to$D     & \textbf{85.2} & \textbf{82.0} & \textbf{80.1} & \textbf{96.0} & \textbf{91.0} & \textbf{89.9} \\
\hline
\end{tabular}
\end{center}
\end{table}

\begin{table}[t]
\centering
\makebox[0pt][c]{\parbox{1\textwidth}{%
\begin{minipage}[b]{0.45\hsize}\centering
\caption{Comparison of different final representations.}
\label{tab:ablation-last-cls}
\resizebox{\linewidth}{!}{
\begin{tabular}{|l|c|c|c||c|c|c|}
\hline
\multirow{2}{*}{Methods} & \multicolumn{3}{|c||}{All (COCO)} & \multicolumn{3}{|c|}{All (VOC 2007)} \\
\cline{2-7}
~ & mAP & OF1 & CF1 & mAP & OF1 & CF1 \\
\hline\hline
Sum         & 84.5 & 81.5 & 79.5 & 94.7 & 89.4 & 88.4 \\
Avg         & 84.5 & 81.5 & 79.2 & 94.8 & 89.6 & 88.6 \\
Max         & 83.9 & 81.2 & 78.8 & 94.7 & 89.6 & 88.8 \\
Bi          & \textbf{85.2} & \textbf{82.0} & \textbf{80.1} & \textbf{96.0} & \textbf{91.0} & \textbf{89.9} \\
\hline
\end{tabular}}
\end{minipage}
\hfill
\begin{minipage}[b]{0.5\hsize}\centering
\caption{Evaluation of activation map generation.}
\label{tab:ablation-spatial-pool}
\resizebox{\linewidth}{!}{
\begin{tabular}{|l|c|c|c||c|c|c|}
\hline
\multirow{2}{*}{Methods} & \multicolumn{3}{|c||}{All (COCO)} & \multicolumn{3}{|c|}{All (VOC 2007)} \\
    \cline{2-7}
    ~           & mAP  & OF1  & CF1  & mAP  & OF1  & CF1  \\
    \hline\hline
    GAP$\to$cls & 85.0 & \textbf{82.0} & 79.8 & 94.8 & 89.7 & 88.5 \\
    GMP$\to$cls & 84.1 & 80.9 & 79.0 & 93.9 & 89.1 & 87.7 \\
    cls $\to$ GMP & \textbf{85.2} & \textbf{82.0} & \textbf{80.1} & \textbf{96.0}   & \textbf{91.0} & \textbf{89.9} \\
    \hline
\end{tabular}}
\end{minipage}
}}
\end{table}
\textbf{Static graph vs Dynamic graph}. We investigate the effects of static graph and dynamic graph in D-GCN. Results are shown in Table~\ref{tab:ablation-combination}. Firstly, we study the case with only one graph. Both static and dynamic graph can achieve better performance compared with baseline ResNet-101, and the dynamic graph performs better on both MS-COCO and VOC 2007. The results show that modeling local (\textit{i.e.}, image-level) category-aware dependencies is more effective than coarse label dependencies over the whole dataset. To further explore whether the static graph is complementary with the dynamic graph, we attempt to combine them in different ways as shown in Table~\ref{tab:ablation-combination}. ``S'' stands for static graph, ``D'' donates dynamic graph, ``P'' denotes that we propagate information through the static graph and dynamic graph in a parallel way, and fuse them by either addition (add) or element-wise multiplication (mul) or concatenation (cat). From the results, ``S$\to$D'' achieves the best performance among all settings. 

\textbf{Final representations}. To demonstrate the effectiveness and rationality of category-specific feature representations, we compare it with image-level feature representations by aggregating the category-specific feature representations to an image feature vector. For aggregation, Sumation (Sum), Average (Avg) and Maximum (Max) are adopted to fuse category-specific feature representations $\textbf{Z}$, which are the output of D-GCN for obtaining image-level feature representations. ``Bi'' means that we utilize binary classifier for each category-specific feature representation to decide whether this class exists or not. Table~\ref{tab:ablation-last-cls} shows the results that the category-specific feature representations outperforms other aggregated representations on all metrics. Thus, we can believe it is an effective way to represent an input image by decomposing the feature map to category-specific representations for multi-label recognition.

\textbf{Evaluation of activation map generation}. As mentioned in Section~\ref{sec:SAM}, we first adopt the standard CAM as baseline. Here we compare the final performance of ADD-GCN between our method and the standard CAM. Specifically, CAM can be donated as ``GAP$\to$cls'' or ``GMP$\to$cls'', and ours is ``cls$\to$GMP''. ``GAP$\to$cls'' equals to ``cls$\to$GAP'' since the classifier is linear operator.
The results are shown in Table~\ref{tab:ablation-spatial-pool}. Comparing  the results of GAP (\textit{i.e.}, GAP$\to$cls) and GMP (\textit{i.e.}, GMP$\to$cls), we believe that GMP loses lots of information as GMP only identify one discriminative part. However, our adaption ``cls$\to$GMP'' outperforms ``GAP$\to$cls'', which indicates that the modified GMP(cls$\to$GMP) may compensate for the disadvantages that ``GMP$\to$cls'' brings.

\subsection{Visualization}
In this section, we visualize some examples of category-specific activation maps and dynamic correlation matrix $\textbf{A}_d$ to illustrate whether SAM can locate semantic targets and what relations dynamic graph has learned, respectively.

    \begin{figure}[t]
    \begin{center}
        \includegraphics[width=0.98\linewidth]{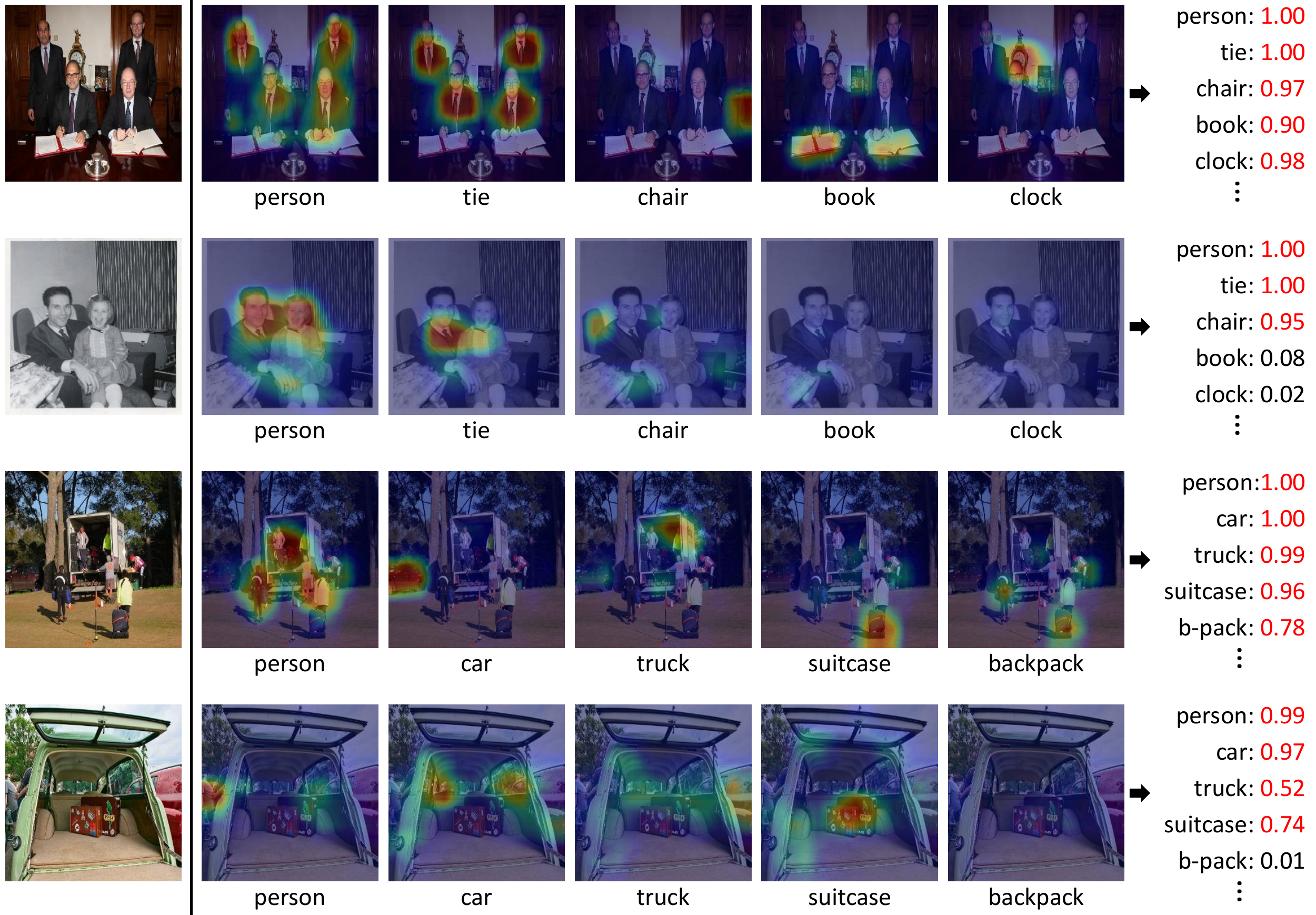}
    \end{center}
    \caption{Visualization results of category-specific activation maps on MS-COCO.}
    \label{fig:visualization}
    \end{figure}
\textbf{Visualization of category-specific activation maps}. We visualize original images with their corresponding category-specific activation maps to illustrate the capability of capturing the semantic region of each category appeared in the image with our SAM module. Some examples are shown in Fig~\ref{fig:visualization}, each row presents the original image, corresponding category-specific activation maps and the final score of each category. For the categories appeared in image, we observe that our model can locate their semantic regions accurately. In contrast, the activation map has low activation of categories that the image does not contain. For example, the second row has labels of ``person'', ``tie'' and ``chair'', our ADD-GCN can accurately highlight related semantic regions of the three classes. Besides, the final scores demonstrate that the category-aware representations are discrminative enough, and can be accurately recognized by our method.

    \begin{figure}[t]
    \centering
    \begin{minipage}{0.34\columnwidth}
        \subfigure[Input Image]{
            \includegraphics[width=1.0\columnwidth]{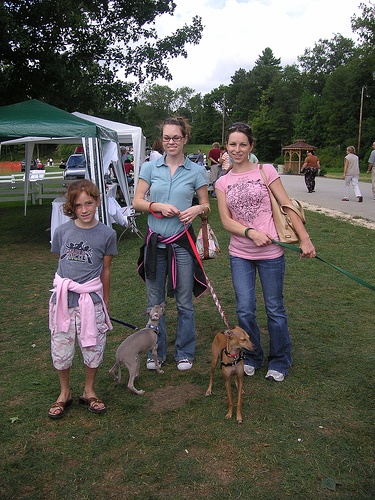}
            \label{fig:dv_inp_img}
        }
    \end{minipage}
    \hfill
    \begin{minipage}{0.61\columnwidth}
        \subfigure[Dynamic Matrix]{
            \includegraphics[width=1\columnwidth]{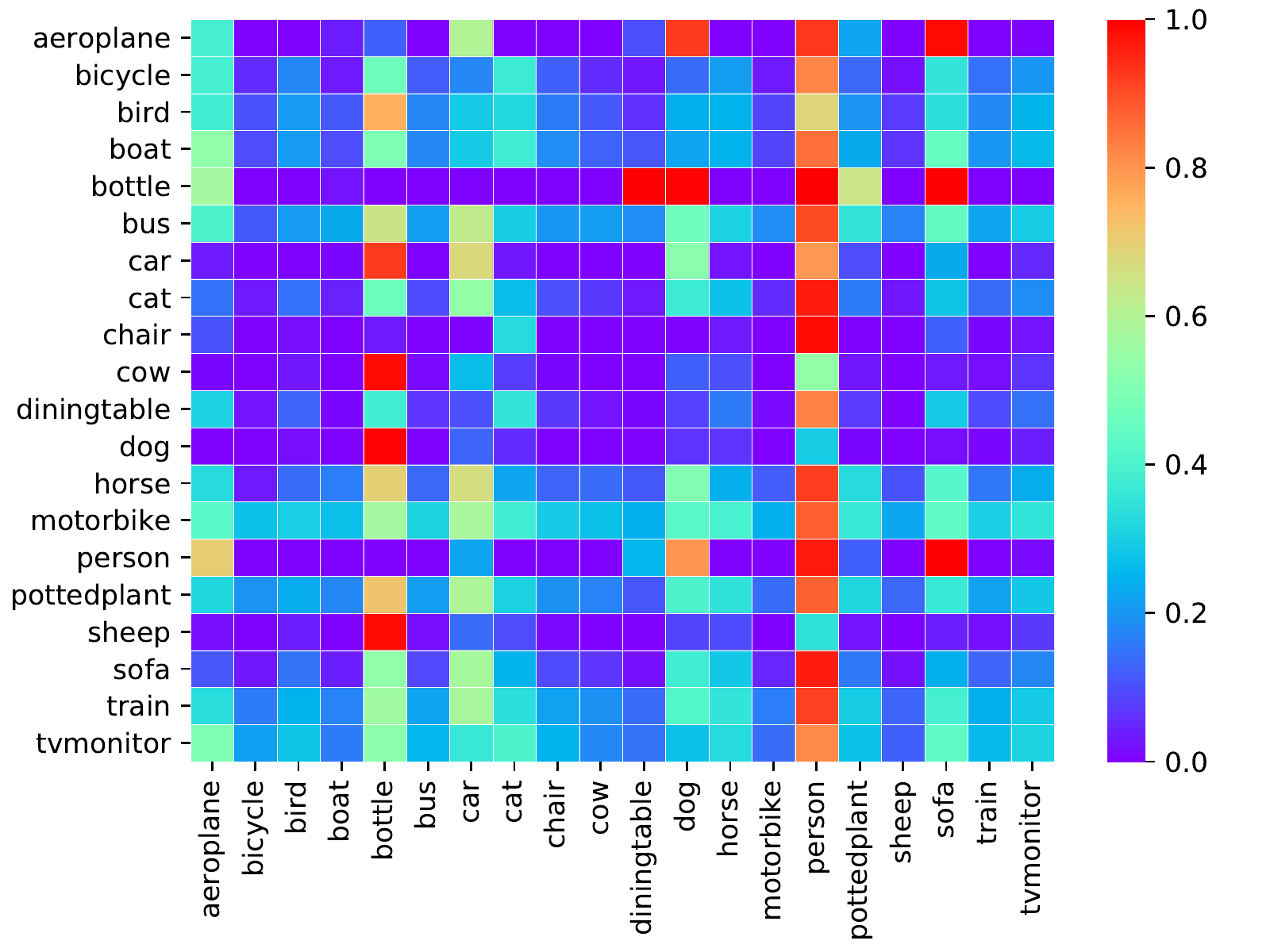}
            \label{fig:dv_dynamic_matrix}
        }\\
    \end{minipage}
    \caption{Visualization of an example and what its dynamic correlation matrix $\textbf{A}_d$ looks like on Pascal-VOC 2007.}
    \label{fig:dynamic_visualization}
    \end{figure} 
\textbf{Visualization of dynamic graph}. As shown in Fig~\ref{fig:dynamic_visualization}, we visualize an original image with its corresponding dynamic correlation matrix $\textbf{A}_d$ to illustrate what relations D-GCN has learned. For the input image in Fig~\ref{fig:dv_inp_img}, its ground truths are “car”, “dog” and “person”. Fig~\ref{fig:dv_dynamic_matrix} is the visualization of the $\textbf{A}_d$ of the input image. We can find that $\textbf{A}_d^{car;dog}$ and $\textbf{A}_d^{car;person}$ rank top~(about top 10\%) in the row of ``car''. It means that ``dog'' and ``person'' are more relevant for ``car'' in the image. Similar results can also be found in the rows of ``dog'' and ``person''. From the observation of the dynamic graph's visualization, we can believe that D-GCN has capacity to capture such semantic relations for a specific input image.

\section{Conclusion}
In this work, we propose an Attention-Driven Dynamic Graph Convolutional Network (ADD-GCN) for multi-label image recognition. ADD-GCN first decomposes the input feature map into category-aware representations by the Semantic Attention Module (SAM), and then models the relations of these representations for final recognition by a novel dynamic GCN which captures content-aware category relations for each image. Extensive experiments on public benchmarks (MS-COCO, Pascal VOC 2007, and Pascal VOC 2012) demonstrate the effectiveness and rationality of our ADD-GCN.

\textbf{Acknowledgements}. This work is partially supported by National Natural Science Foundation of China (U1813218, U1713208), Science and Technology Service Network Initiative of Chinese Academy of Sciences (KFJ-STS-QYZX-092), Guangdong Special Support Program (2016TX03X276), and  Shenzhen Basic Research Program (JSGG20180507182100698, CXB201104220032A), Shenzhen Institute of Artificial Intelligence and Robotics for Society. We also appreciate Xiaoping Lai and Hao Xing from VIPShop Inc. who cooperate this project with us and provide validation Fashion data. 



%
%
\bibliographystyle{splncs04}
\bibliography{egbib}
\end{document}